\documentclass[sigconf]{acmart}

\usepackage{booktabs} 
\usepackage{makecell}
\setcellgapes{4pt}
\usepackage[ruled]{algorithm2e} 
\usepackage{url}
\usepackage{amsmath}
\usepackage{algorithmic}
\usepackage{amssymb}
\usepackage{siunitx}
\usepackage{gensymb} 
\usepackage{graphicx}

\usepackage{array, multirow, caption} 

\setcopyright{rightsretained}

\acmDOI{10.475/123_4}

\acmISBN{123-4567-24-567/08/06}

\acmConference[Shenzhen]{ENSsys}{Nov 2018}{}
\acmYear{1997}
\copyrightyear{2016}

\newcolumntype{C}{ >{\centering\arraybackslash} m{4cm} }
\newcolumntype{D}{ >{\centering\arraybackslash} m{1cm} }

\acmArticle{4}
\acmPrice{15.00}

\editor{Jennifer B. Sartor}
\editor{Theo D'Hondt}
\editor{Wolfgang De Meuter}

\hypersetup{draft} 
\begin{document}

\title{Scaling Configuration of Energy Harvesting Sensors with Reinforcement Learning}	

\author{Francesco Fraternali}
\orcid{1234-5678-9012-3456}
\affiliation{%
  \institution{University of California, San Diego}
  }
  \email{frfrater@ucsd.edu}
\author{Bharathan Balaji}
\affiliation{%
  \institution{University of California, Los Angeles}
}
\email{bbalaji@ucla.edu }

\author{Rajesh Gupta}
\affiliation{%
  \institution{University of California, San Diego}
}
\email{gupta@eng.ucsd.edu }

\renewcommand{\shortauthors}{Fraternali et al.}

\begin{abstract}
With the advent of the Internet of Things (IoT), an increasing number of energy harvesting methods are being used to supplement or supplant battery based sensors. Energy harvesting sensors need to be configured according to the application, hardware, and environmental conditions to maximize their usefulness. As of today, the configuration of sensors is either manual or heuristics based, requiring valuable domain expertise. Reinforcement learning (RL) is a promising approach to automate configuration and efficiently scale IoT deployments, but it is not yet adopted in practice. 
We propose solutions to bridge this gap: reduce the training phase of RL so that nodes are operational within a short time after deployment and reduce the computational requirements to scale to large deployments. We focus on configuration of the sampling rate of indoor solar panel based energy harvesting sensors. We created a simulator based on 3 months of data collected from 5 sensor nodes subject to different lighting conditions. Our simulation results show that RL can effectively learn energy availability patterns and configure the sampling rate of the sensor nodes to maximize the sensing data while ensuring that energy storage is not depleted. The nodes can be operational within the first day by using our methods.
We show that it is possible to reduce the number of RL policies by using a single policy for nodes that share similar lighting conditions.
\end{abstract}

%
%

\ccsdesc[500]{Computing methodologies~Reinforcement learning}
\ccsdesc[500]{Computer systems organization~Sensor networks}

\keywords{Internet of Things, Reinforcement Learning, Battery-Less, Scaling}

\acmYear{2018}\copyrightyear{2018}
\setcopyright{acmcopyright}
\acmConference[ENSsys '18]{ENSsys '18: International Workshop on Energy Harvesting $\$$0 Energy-Neutral Sensing Systems}{November 4, 2018}{Shenzhen, China}
\acmBooktitle{ENSsys '18: International Workshop on Energy Harvesting $\$$0 Energy-Neutral Sensing Systems, November 4, 2018, Shenzhen, China}
\acmPrice{15.00}
\acmDOI{10.1145/3279755.3279760}
\acmISBN{978-1-4503-6047-0/18/11}

\maketitle

\section{Introduction}
The number of connected Internet of Things (IoT) devices is expected to increase from 27 billion in 2017 to 125 billion in 2030~\cite{link:iot}. Driven by the need to collect data about the environment and human behavior, an increasing number of applications have emerged and produce useful, scientifically-relevant data.
Energy harvesting sensors are a key part of the IoT ecosystem to reduce dependence on batteries \cite{paper:IoTeverywhere,Paper:Flicker:Hester:2017:FRP:3131672.3131674, renner2013sustained,Hester:2017:TEI:3131672.3131673}. However, the performance of energy harvesting sensors is highly dependent on energy availability in the environment~\cite{paper:campbell_cinamin,paper:campbell_energy_architecture_building, Pible}. To maximize performance, the sensors need to be tuned carefully according to the application requirements, hardware capabilities and environmental conditions. As IoT deployments scale, setting these configuration parameters manually or based on heuristics becomes infeasible. 
We explore Reinforcement Learning (RL) as a promising solution as it can dynamically set these parameters by online learning. 

We focus on indoor solar energy harvesting sensors and use RL to dynamically configure the sampling rate of the sensor. If the sampling rate is too high, the node expends available energy and reduces its uptime. If the rate is too low, it impacts the application performance. The objective is to configure the sensing period such that the number of sensor samples is maximized while ensuring that the sensor node does not run out of energy.

Several solutions propose machine learning techniques to automatically configure sensors~\citep{paper:RL-adapting,paper:RL-ener-harvest-6797906, paper:RL-ener-harvest-QoS-5284227}, but their results are limited:
\textit{(i)} they are based on short simulations and do not capture pragmatic aspects of a real-life environment; \textit{(ii)} do not consider the scaling of the system to thousands of nodes.
To overcome these limitations we conduct a series of simulation experiments to evaluate solutions for auto-configuration of sensors using RL. The domain expert specifies the parameters (i.e. actions in RL terminology) to be configured, the contextual features (i.e. states) which affect these parameters and the utility (i.e. reward) that they seek to maximize. RL then tries out different parameter values, observes its effect on the given utility and learns the optimal sensor-node configuration to maximize the long-term utility according to the environmental situation.

We built a generic sensor node with a solar panel for energy harvesting, supercapacitor for energy storage, BLE radio for communication and general sensors such as light and temperature\cite{Pible}. 
We deployed five nodes in different indoor lighting conditions collecting light intensity and supercapacitor voltage level for 3 months in our department building. 
We developed a simulator based on this data that models the essential aspects of the sensor nodes and its environment and train different RL agents using the Q-learning algorithm~\cite{watkins1989learning}. We show how the system adapts to environmental changes and appropriately configures each mote to maximize sensing-rate while avoiding energy storage depletion. 

Typical RL solutions need significant historical data or an online training phase where they explore the solution space randomly to effectively learn a strategy. However, this is detrimental to sensor deployments as historical data is expensive to collect, and a long training phase makes the sensor node unusable immediately after deployment. To combat this, we propose an adaptive on-policy RL solution that reduces the training phase after deployment. We show that nodes can effectively operate, i.e. sense data periodically without depleting the stored energy, within the first day. We also show that similar results can be obtained by exploiting transfer learning. Finally, prior solutions consider one RL policy for each sensor node and affect the scalability. We show that it is possible to use a single policy for sensors that share similar lighting conditions and still effectively configure the sensor nodes. 
\vspace{-1mm}
\section{Related Work}
\label{RelWork}
The importance of automatic sensor configuration to reduce manual intervention is underlined by many works~\cite{paper:IoT-reconfigurable,paper:direct-RL, paper:sens-average,paper:RL-adapting,paper:RL-energy-harvesting,paper:RL-ener-harvest-QoS-5284227}. Due to the close relationship between data quality and energy consumption~\cite{paper:data-energy-tradeoff}, a sensor-node should adapt its sensing to meet application requirements while avoiding energy depletion~\cite{paper:iot-powering}.
Prior works have proposed adaptive duty cycling on energy harvesting sensors to achieve energy neutral operations \cite{paper:Mani-power-management, hsu2006adaptive, paper:sens-average}. Based on the predicted energy, nodes adjust their duty-cycle parameters and increase lifetime and applications performance \cite{sudevalayam2011energy}. Like prior work, RL uses prediction to make decisions, but unlike prior work the policy is learned automatically to converge to the optimal solution.

Machine learning techniques have been adopted to predict the future energy availability of a sensor node and select the correct sensor parameter configuration~\cite{paper:RL-review}. For example, Dalamagkidis et al.~\cite{paper:RL-building} and Udenze et al.~\cite{paper:direct-RL} show that Reinforcement Learning (RL) outperforms traditional on/off controller and a Fuzzy-PD controller. RL has been widely adopted to improve wireless sensor network performance: to dynamically select at run time a routing protocol from a pre-defined set of routing options, which provides the best performance \cite{paper:RL-mesh}; to bring wireless nodes to the lowest possible transmission power level and, in turn, to respect the quality requirements of the overall network \cite{paper:RL-power-chincoli2018self}; to adapt sampling intervals \cite{paper:RL-adapting} in changing environments. 
Overall, RL promises to learn the optimal policy that is specific to each context and application. Hence, it helps push the boundaries of what is possible in sensor networks. RL can be implemented local to the device as well, the Q-table does not take up much memory or compute.
However, prior works in RL based sensor configuration are not considering many aspects of the design required for a large-scale real-world deployments of thousands of nodes.

Simulation results by Dias et al.~\cite{paper:RL-adapting} optimize energy efficiency based on data collected during five days by five sensor nodes. The data collection period is just too short to capture wide range of real environmental changing conditions. Moreover, they assume a fixed 12 hours period as the time needed by the Q-Learning algorithm to "calibrate" the action-value function for the rest of the 4.5 days experiment. But a fixed Q-Learning training time can not capture all kind of environmental changes: faster environmental changes could require higher calibration time, while slow environmental changes could relax. In this paper, we propose a dynamic on-policy training interval, that dynamically varies the time between trainings. On a system that includes thousand of nodes, the time between two consecutive on-policy training is important because can reduce computation and cost. 
Similarly, Yue et al.~\cite{paper:RL-energy-harvesting}, show a dynamic power management method for increasing battery life of mobile phones. Although they use realistic simulation models for battery use, they use data from Linux network trace for simulations that does not capture environmental conditions of the target application. 
RL has also been used to improve the energy utilization for energy harvesting wireless sensor networks \cite{paper:RL-ener-harvest-QoS-5284227,paper:RL-ener-harvest-6797906}. Hsu et al.~\cite{paper:RL-ener-harvest-6797906} apply RL for sustaining perpetual operation and satisfying the throughput demand requirements for today's energy harvesting wireless sensor nodes. However, their algorithm is built and tested in an outdoor environment, where sunlight patterns are consistent throughout the day (i.e., light is available from sunrise to sunset). In our work, we focus on indoor sensing where the daylight patterns are less well-defined and the light availability is also affected by human occupancy. Furthermore, we focus on learning faster and scaling better with (1) adaptive policy learning (2) transfer RL (3) using common policy for nodes in similar environments.

\vspace{-1mm}
\section{Problem Formulation}
\label{Design}


\subsection{Problem Statement}
Nodes can be placed in different locations in the building. Each sensor node will be subject to different light patterns that are determined by human behavior (i.e. lights turned on and off) or by natural light. Light availability will vary from weekdays to weekends, from winter to summer and with changes in usage patterns, e.g. a conference room vs a lobby. 
A solar panel based energy harvesting node needs to automatically adapt its sampling rate to these changing conditions so as to maximize the utility to its applications, i.e., maximize sensor sampling while keeping the node alive. 
We use sensing frequency as a measure of QoS, but RL can be adapted to other metrics by changing the reward function.
\vspace{-1mm}
\subsection{Reinforcement Learning and Q-Learning}
\label{RL}

In a typical Reinforcement learning (RL) problem \cite{sutton1998reinforcement}, an agent starts in a state $s$ and by choosing an action $a$, it receives a reward $r$ and moves to a new state $s^\prime$. RL agent's goal is to find the best sequence of actions that maximizes the cumulative long term reward. 
The way the agent chooses actions in each state is called its policy.
\begin{equation}
s \overset{a}{\underset{\pi}\rightarrow} r, s^\prime
\end{equation}
For each given state $s$ and action $a$, we define a function $Q(s, a)$ that returns an estimate of expected total reward by starting at state $s$, taking the action $a$ and then following a given policy $\pi$. $Q^*$ is the $Q$ value obtained using the optimal policy $\pi^*$ that maximizes its expected cumulative long term reward.
\begin{equation}
\label{eq:recursive}
Q(s, a) = r_0 + \gamma r_1 + \gamma^2 r_2 + \gamma^3 r_3 + ... 
\end{equation}
where $\gamma\in[0,1]$ is called a discount factor and it determines how much the function $Q$ in state $s$ depends on the future actions, the rewards exponentially diminish the further they are in the future. 
In Q-learning, the policy picks the action that has the highest Q-value in each state. Thus, we obtain the classic Q-function~\cite{watkins1989learning}:
\begin{equation}
Q(s, a) = r + \gamma max_a Q(s', a)
\end{equation}
The algorithm starts with a randomly initialized Q-value for each state-action pair and an initial state $s_0$. An episode is defined as a sequence of state transitions from the initial state to the terminal state. The Q-learning algorithm visits each state-action pair multiple times in an episode. It follows a $\epsilon$-greedy policy, where for each state it picks the action that has the maximum Q-value with probability $(1-\epsilon)$ and a random action otherwise. The reward obtained by selecting the action is used to update the Q-value with a small learning rate. Under the conditions that each of the state-action pairs are visited infinitely often, the Q-learning algorithm is proven to converge to the desired $Q^*$~\cite{watkins1992q}. 

\begin{figure}[ht]
\centering
     \includegraphics[width=0.9\linewidth]{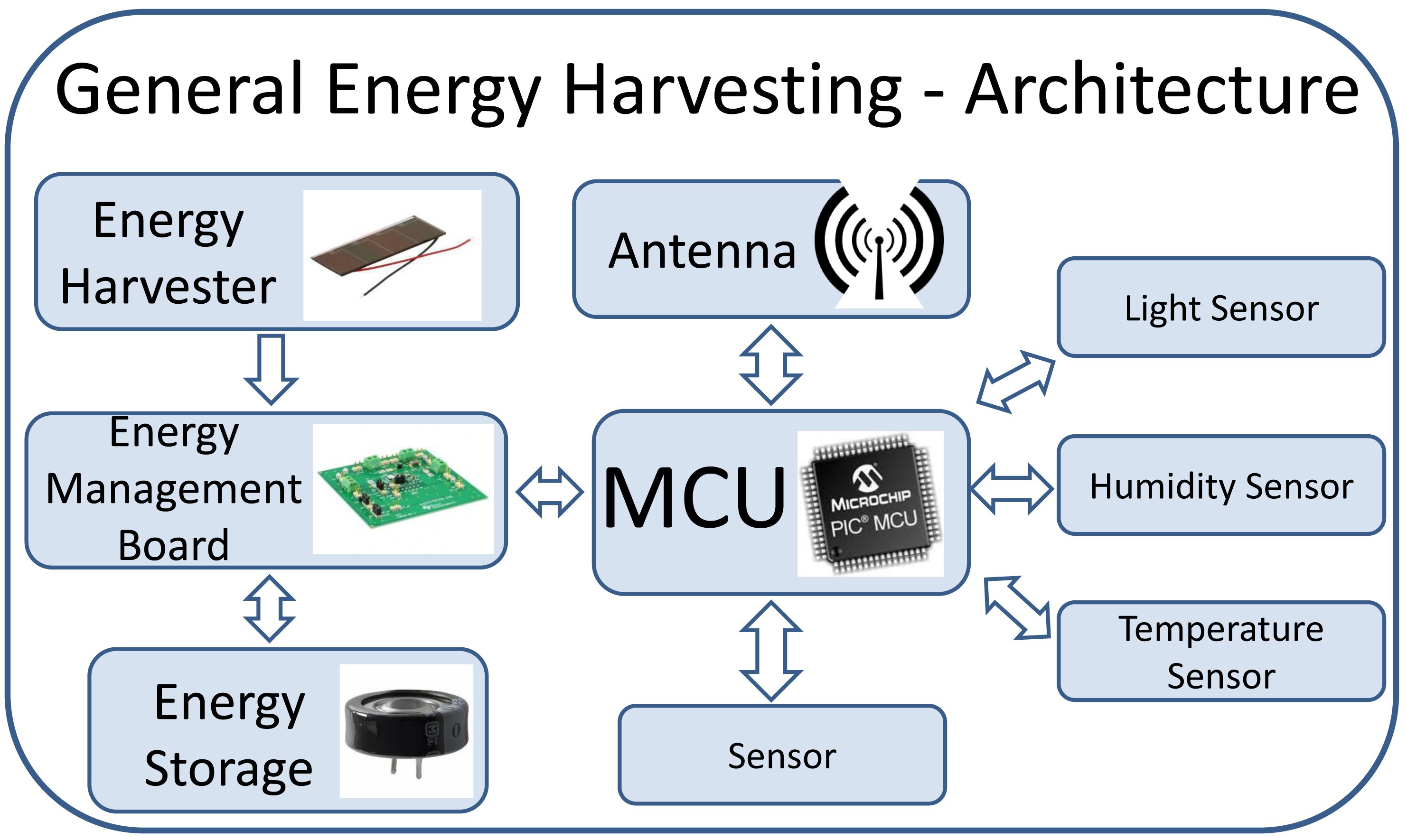}
     \vspace{-1em}
    \caption{Energy Harvesting Node Block Diagram. 
    }
    \vspace{-1em}
    \label{fig:Components}
\end{figure}

\vspace{-1mm}
\subsection{RL for Configuring Sensors}
\begin{figure}
  \centering
     \includegraphics[width=0.9\linewidth]
{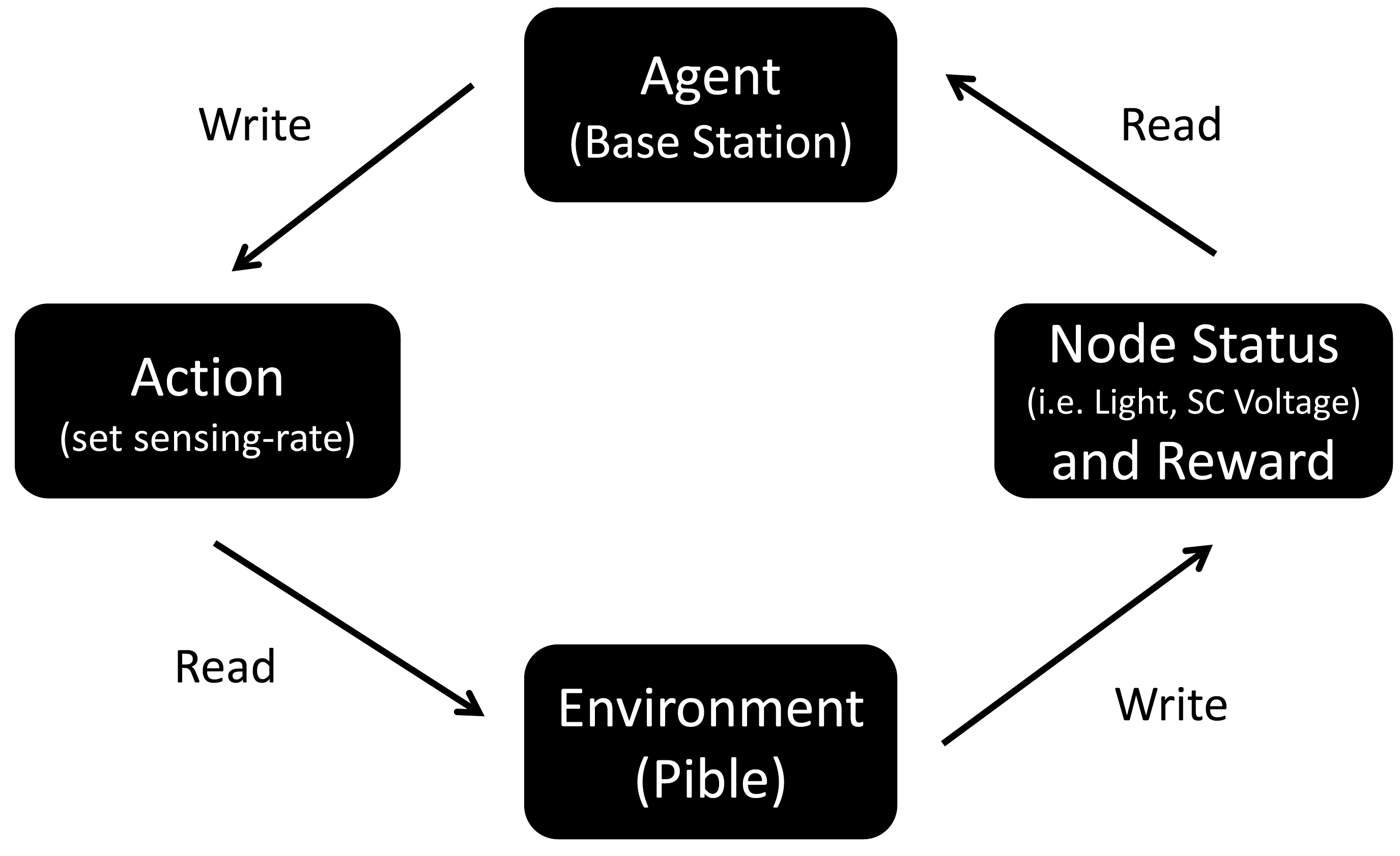}
\vspace{-1em}
    \caption{RL Communication Process: Block Diagram.} 
		\vspace{-1em}
    \label{fig:Circle_Life}
\end{figure}

Figure \ref{fig:Circle_Life} shows the overview of our approach. Our battery-less sensor node (i.e. Pible) collects energy from a solar panel and sends sensor data to a base station. 
The base station determines the sampling rate to set based on data provided by the node.



\textbf{Agent:} The agent is the brain of our system and learns from the environment by collecting data from the sensor nodes and by communicating back to them what action they need to take next. In our setting, the agent resides in the base station that the sensor node communicates with and the compute intensive RL training is done at the base station level. 

\textbf{Environment:} 
It is represented by the sensor nodes (i.e. Pible) that sense the environment and communicate information back to the agent. Data sent includes light intensity, voltage level of the energy-storage element, and the current performance state (i.e. sampling rate).
Super-capacitor voltage data are used to track energy consumption trends and calibrate simulation parameters.


\begin{table}[h]
\centering
  \caption{Sensing Rate based on Performance State}
  \label{tab:Perf-State}
  \begin{tabular}{cc}
    \toprule
    Performance State & Sensing Rate [s]  \\
    \midrule
    3 & 15\\
    2 & 60 (1 min)\\
    1 & 300 (5 min)\\
    0 & 900 (15 min)\\
  \bottomrule
\end{tabular}
\end{table}

\textbf{Actions:} Typical building sensors use sensing rate in the order of minutes, ranging from tens of seconds to 1 hour \cite{paper:augment_audits-finnigan2017augmenting, link:sens-rate-1,link:sens-rate-2}. 
Hence, we discretize the sensing rate to four performance states as reported in Table \ref{tab:Perf-State}. The action selected corresponds to the performance state to use, e.g. taking action 2 corresponds sending data every minute. The discretization reduces the action space, and hence, decreases the convergence time of the Q-learning algorithm. 

\textbf{State:} The state is used by the agent to pick the actions. We use three states for our system: \textit{(i)} light intensity level, \textit{(ii)} energy storage level, \textit{(iii)} weekend/weekdays. We discretize light intensity and energy storage levels to 10 values each.

\textbf{Reward Function:} 
The goal of the system is twofold: 
\textit{(i)} Send as much data as possible by increasing performance states. 
\textit{(ii)} Avoid energy storage depletion. 
The reward function decides the trade-off between the two factors. 
\begin{itemize}
\item[--] Reward = Performance States (i.e. 0, 1, 2 or 3)

\item[--] Reward = -300 if energy storage level reaches 0 (node dies).

\end{itemize}


The reward value of -300 has been calculated considering that we use 24 hours as our horizon 
 and the RL agent selects an action every 15 mins. The accumulated rewards by sensing rapidly cannot exceed the energy depletion penalty. 

\textbf{State Transitions:} The agent observes state transitions and takes actions, i.e. sends commands to change the sampling rate, every 15 minutes. 

\label{sec:rl_circle}
\begin{algorithm}[ht]
  \caption{RL Algorithm for Energy Harvesting Sensors}
  \label{algoRL}
  \begin{algorithmic}[1]
  \STATE Initialize q$_{table}$ as an empty set
  \STATE Initialize control action a, state s$_{curr}$ and s$_{next}$
  \STATE $\epsilon_{max}$ = 1
  \STATE $s_{curr}$ $\leftarrow$ Sense Environment 
  \WHILE{time passed $<$ 24 hours}
  
  \STATE \newenvironment{rcases}
  {\left\lbrace\begin{aligned}}
  {\end{aligned}\right.}

\begin{equation*}
\text{ a = }
\begin{rcases}
  \text{if (random(n)} > \epsilon \text{)   take:   }  argmaxQ(s_{curr}, a')\\
  \text{otherwise:    }  \text{    take a random action}
\end{rcases}
\end{equation*}
\STATE wait 15 mins
\STATE $s_{next}$ $\leftarrow$ Sense Environment 

\STATE r = reward(s$_{curr}$, a, s$_{next}$)

\STATE $ q_{predict} = q_{table}[s_{curr}, a] $
\STATE $ q_{target} = r + \gamma * max(q_{table}[s_{next}, all actions]) $
\STATE $ q_{table}[s_{curr}, a] += \alpha * (q_{target} - q_{predict}) $

\STATE $ \epsilon_{max} = max(\epsilon_{max} - \Delta, \epsilon_{min}) $
\STATE $ s_{curr} \leftarrow s_{next} $
  \ENDWHILE
  \end{algorithmic}
\end{algorithm}


A limitation of the proposed method is that it is not robust to communication failures (i.e. base station and sensor node communication). But the Q-learning policy can be executed inside the node to make it robust. The memory and compute requirements are low.

\vspace{-1mm}
\subsection{RL Algorithm}
Algorithm \ref{algoRL} lays out our Q-learning steps. After initializing the Q-table to a matrix of zeros and setting $\epsilon$ to the maximum value (line 1-3), the agent receives the node status (i.e. light, energy storage voltage and performance state) on line 4. At this point, the algorithm enters a while loop that lasts for an episode (i.e. 24 hours) and starts by selecting an action (line 6). The action is selected by following the $\epsilon$-greedy policy. 
At the beginning, $\epsilon$ is initialized to 1, the algorithm selects a lot of random actions (i.e. exploration) in the first phase of the learning. But for each executed while loop, $\epsilon$ \emph{is decreased} by $\Delta$ (line 13), causing the action policy to select more exploitive actions over time.
Based on the action selected, the environment changes its status and sends the updated state to the agent after 15 minutes (line 7-8). A reward is given to the agent based on the current status, action and next node status (line 9). 
The algorithm calculates the difference between q$_{predict}$ and q$_{target}$ where q$_{predict}$ is the value in the Q-table for the state-action pair and q$_{target}$ is the the reward obtained by taking action $a$ plus the rewards obtained by picking the actions with the highest Q-value until the end of the episode. Their difference is scaled by a learning factor $\alpha$ and added to the Q-value of the corresponding state-action pair. 
Finally, on line 14 we save the value of the next state to the current state and repeat the while loop until the end of an episode. 
We assume our system has reached convergence when the mean of all Q-Table's value does not change over time. 
We want to underline that Q-learning learns the probability of transition from one state (light, voltage levels) to another. Hence, it will be robust to spurious changes to voltage levels as long as it doesn't happen consistently.

\subsection{Sensor Node}
As a target, we used our energy harvesting platform build for general building applications \cite{Pible}.
Pible's energy harvesting sensor architecture is depicted in Figure~\ref{fig:Components}: a solar panel and transfers power to an energy management board, which stores the accumulated energy in a supercapacitor. Once the energy accumulated reaches a usable voltage level, the energy management board powers the micro-controller (MCU) that starts its operations.

We made further improvements to the hardware design to increase operational time without light. We adopt a 1F super-capacitor with a higher nominal voltage (i.e. 5.5V) to store more energy (E = 0.5*V*V*C). To allow the MCU to read this high voltage, we introduced a voltage divider that uses very high resistor value (10 MOhm) to minimize leakage current. With these improvements, the Pible node achieves up to a week of lifetime without light by sensing one sensor every 10 minutes. For this work, we consider the same power consumption measurements as in \cite{Pible}.


\section{Scaling Methods}
\subsection{Day-by-Day Learning - Baseline}
\label{baseline}
As a baseline, we learn a new Q-table daily based on light data collected each day~\cite{paper:RL-ener-harvest-QoS-5284227,paper:RL-ener-harvest-6797906}. The sensor node starts with a fixed policy on the first day, and collects data every 5 min. We use the collected data in our simulator to train the RL agent using Q-learning. The learned Q-table based policy is used to collect the data the next day. At the end of the day, the collected data is again used to create a new Q-table via Q-learning. Instead of initializing the Q-table to zero, the training starts from the previous day's Q-table that summarizes the learning until that day. The learned Q-table is used as the deployment policy for the next day and the cycle continues. 
\vspace{-2mm}
\subsection{Dynamic Policy Training Interval} 
The day-by-day learning trains a new policy every day. Instead of a static update interval, we propose a dynamic approach, where the policy updates happen over shorter interval initially when the agent is learning and the interval is increased as learning stabilizes. If environmental conditions change, the interval can again be shortened to encourage faster learning.
Starting from an empty Q-Table, we run on-policy training simulations every hour. As soon as the Q-Table converges, we use this new generated Q-Table to run a simulation that calculates the total reward based on the light-data collected so far. If the reward results are equal or better compared to the reward results obtained with the old Q-Table, we double the time we run the on-policy training. At the contrary, if the total reward achieved by the new Q-Table is less w.r.t the old Q-Table, we halve the on-policy training interval time up to a minimum of 1 hour. We show that this simple adaptive strategy significantly speeds up training. The halving and doubling of interval is itself a heuristic, but can be generalized with meta learning methods~\cite{finn2017model}.
\vspace{-2mm}
\subsection{Sharing RL Policy} 
Prior works~\cite{paper:RL-adapting,paper:RL-ener-harvest-6797906,paper:RL-ener-harvest-QoS-5284227} and the methods proposed thus far use a different RL policy for each sensor node. As we scale to thousands of nodes, policy training becomes infeasible in embedded computers such as Raspberry Pi typically used as the base station. We can potentially just use a single policy across all the sensor nodes, but our experiments show that the performance of the nodes drop significantly when using a common policy as the environmental conditions that each sensor is exposed to is vastly different. We propose using a single policy between the sensor nodes that share similar energy availability. The sensor nodes can be clustered based on lighting data collected. While we leave the clustering method to future work, we show in simulation that the nodes perform satisfactorily when they use a single policy for hundreds of sensor nodes that share lighting characteristics. 
\vspace{-2mm}
\subsection{Transfer Learning} 
We study the effect of learning a new RL policy by exploiting a pre-learned Q-Table instead of learning everything from scratch~\cite{taylor2009transfer}. This is important to speed up the learning process of the nodes. 

\section{Simulation Results}
We build our simulator using real power consumption data. The base station uses BLE gattool functions to exchange data with the sensor nodes. As soon as a sensor-node node wakes-up, it starts advertising. The base station reads the advertisement, connects to the sensor-node and exchanges data. The energy consumption of this communication process is taken into account in our simulation. The whole process lasts a few seconds and it does not impact the performance of the system.
Before running the simulations, we calibrate the simulation's parameters using real discharging measurements. To simulate charging, we used a linear model considering PV cells information from datasheets \cite{link:sanyo}.

\subsection{Simulation Setup}
\label{sec:sim_sett}
\subsubsection{Modeling the States for RL}
\begin{itemize}
\item[--] \textit{Light intensity level:} we normalize the light intensity from a range of 0 to 10, where 0 represents no light and 10 represents 2000 lux. We select 2000 lux as a maximum value after checking typical indoor light intensity in buildings, values above 2000 lux are approximated to 10. 

\item[--] \textit{Energy storage level:} we scale the energy storage voltage calculated to a value from 0 (i.e. minimum voltage available of 2.1V) to 10 (i.e. max voltage available of 5.5V).

\item[--] \textit{Weekend/Weekday:} buildings indoor lights patterns are strongly dependent on the presence of people \cite{paper:campbell_cinamin}. Hence, we consider a binary state to capture weekdays and weekends. 



\end{itemize}


\subsubsection{RL Hyper-parameters}

Table \ref{tab:HyperP-Simu} reports the hyper-parameters used for the Q-learning algorithm for our simulations.

\begin{table}[ht]
\centering
\vspace{-2mm}
  \caption{Simulation RL Hyper-parameters}
  \label{tab:HyperP-Simu}
  \begin{tabular}{c | c c c c c}
    \toprule
    Hyper& Reward & Epsilon & Epsilon & Epsilon & Learn\\
   Parameter & decay & max & min & decrement & rate \\
  & ($\gamma$) & ($\epsilon_{max}$)& ($\epsilon_{min}$)  & ($\Delta$) & ($\alpha$)\\
    \midrule
    Value   & 0.99 & 1 & 0.1 & 0.0004 & 0.1 \\
  \bottomrule
\end{tabular}
\end{table}

\subsection{Day-by-Day Learning - Baseline Results}
\label{sec_baseline_results}

\begin{figure}[th]
	\centering
	\includegraphics[width=\linewidth]{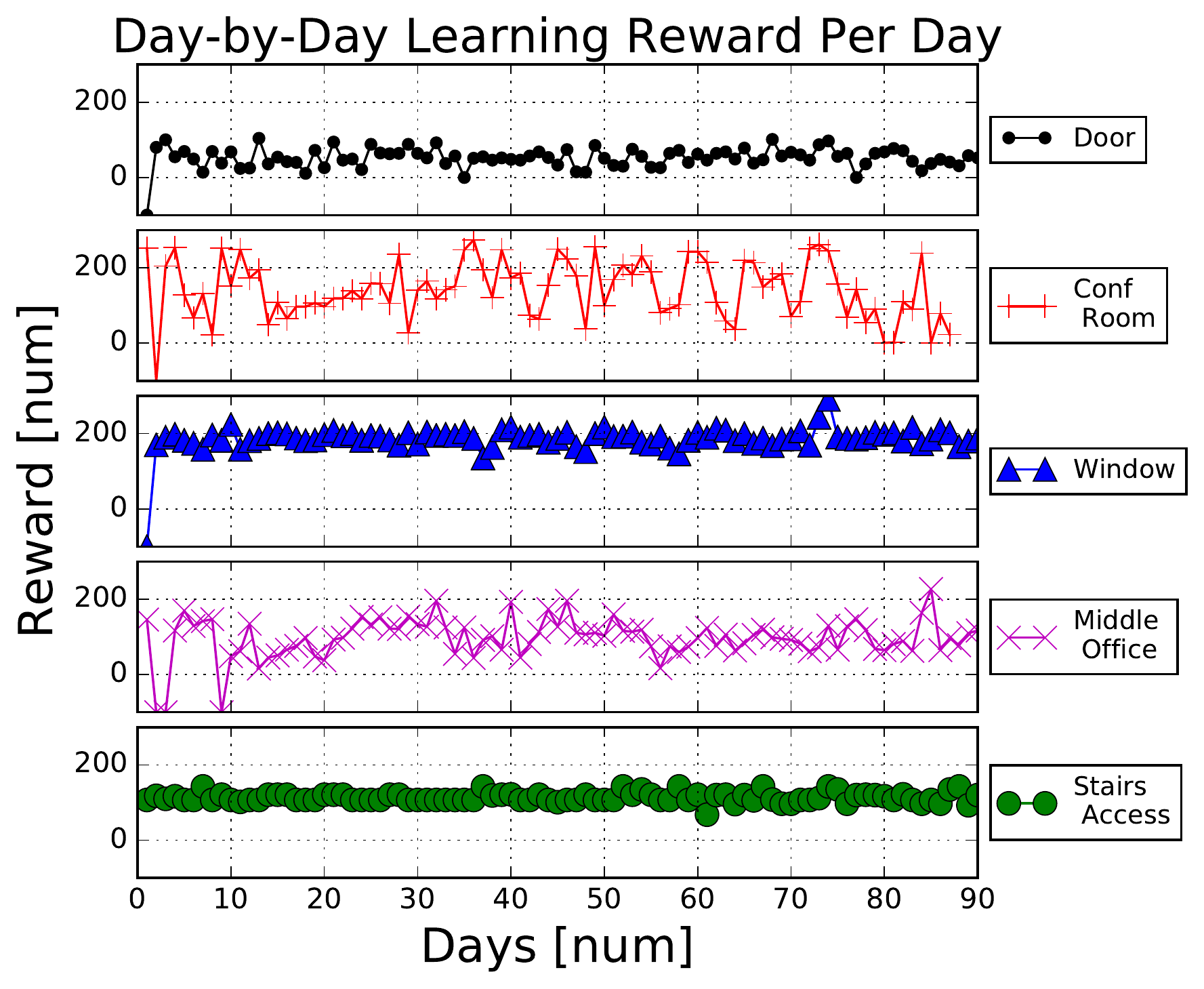}
	\vspace{-2mm}
	\caption{Simulation Results on Different Lighting Conditions using Day-by-Day Learning (i.e. our Baseline)}
	\label{fig:DbD}
    \vspace{-1em}
\end{figure}

We run 5 different simulations, one for each node in a different lighting condition. 
Figure \ref{fig:DbD} shows the results, each dot represents the reward achieved by the node at the end of a day.

Several nodes (i.e. Door, Window, Conference Room and Middle of Office) are receiving negative rewards (that we report as -100 to improve graph visibility) on the first few days indicating that the Q-learning algorithm needs time to learn and adapt to changes in the environment. Furthermore, we can notice that the nodes that are receiving negative rewards even after the RL started is learning are the one subject to human-dependent light pattern such as \textit{Conference Room} (that has no windows and light is only on when people enter the room) and \textit{Middle of an Office}. On the other hand, nodes that are subject to constant light patterns such as \textit{Windows} (that is subject to ambient light from sunrise to sunset), \textit{Door} (that has a window in front of it) limit the days in which the energy storage is depleted just on the first day. Finally, the \textit{Stair Access} case, where light is always on for security reasons, receives always positive reward due to the stability of its light pattern.
Overall, we can notice that after a week, all the light patterns are learned by the system and the rewards follow a constant pattern that depend on the light availability of each node placement. On average the nodes are operational within the first 2.6 days of deployment when training from scratch.

\subsection{Dynamic Policy Training Interval Results}

\begin{figure}[th]
	\centering
	\includegraphics[width=0.9\linewidth]{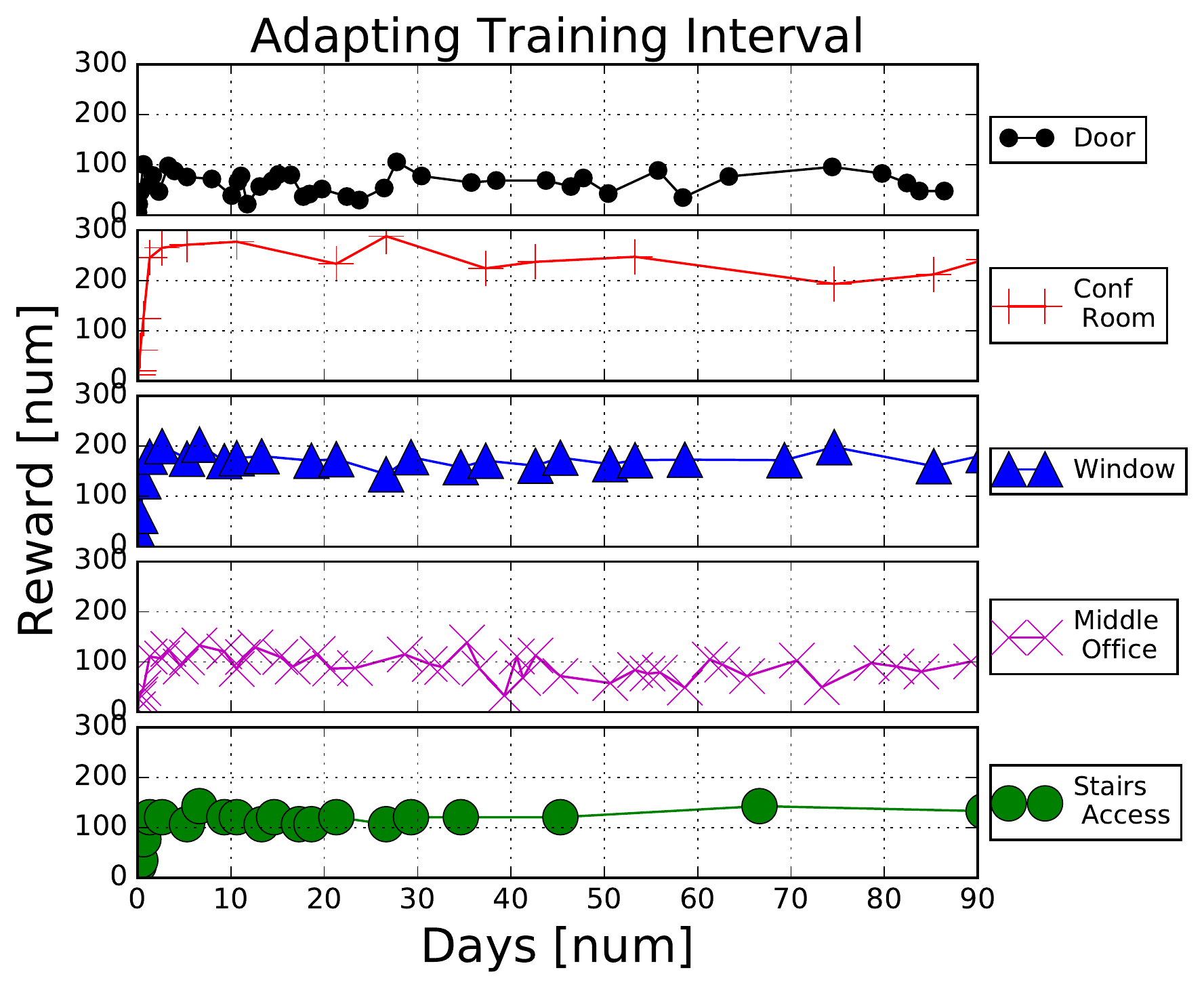}
     \vspace{-4mm}
	\caption{Dynamic Policy Training Interval Results}
    \label{fig:auto_int}
    \vspace{-1mm}
\end{figure}

Figure \ref{fig:auto_int} reports the results of our experiment. During the first week of training, the nodes use a short on-policy training interval, confirming that they are still exploring the environment. But after 10 days, they are able to drastically reduce the number of on-policy training performed. Furthermore, all the nodes were able to maintain positive rewards and avoid energy storage depletion. 

\begin{table}[ht]
\centering
\vspace{-2mm}
  \caption{Comparison between the dynamic policy training interval experiment w.r.t. day-by-day approach: number of policy trainings performed and number of days the nodes receive a negative reward (between parenthesis)}
  \label{tab:dynamic_res}
  \begin{tabular}{c c c c c c}
    \toprule
    Num Training & Window & Middle & Door & Confer & Stair\\
   (Num Nod Died) &  & Office &  & Room & Access\\
    \midrule
    Baseline & 89 (1) & 89 (3) & 89 (1) & 89 (2) & 89 (0)  \\\hline
    
    Dynamic & \multirow{2}{0.65cm}{27(0)} & \multirow{2}{0.65cm}{42(0)} & \multirow{2}{0.65cm}{41(0)}&\multirow{2}{0.65cm}{17(0)} & \multirow{2}{0.65cm}{20(0)} \\
    On-Policy&  &  &  &  & \\\hline
  Percentage  & 69 \% & 53 \% & 54 \% & 81 \% & 77 \% \\
  \bottomrule
  \vspace{-1em}
\end{tabular}
\end{table}

Table \ref{tab:dynamic_res} compares the dynamic training approach with the day-by-day training quantitatively. 
The number of On-Policy training is drastically reduced by using our method: on the \textit{Conference Room} case the number of On-Policy training can be reduced up to a 81\% while in the worst case (i.e. \textit{Middle Office}) we can reduce it to up 53\%. In the same Table, we reported between parenthesis the number of days in which the system achieves negative rewards and we can notice that our dynamic On-Line Policy remove all the negative rewards accumulated on the first days by the Baseline approach and the nodes become operational within the first day. 

\subsection{Sharing RL Policy Results}
\label{exp_scaling}
We exploit the 5 real data traces from our sensor nodes across 5 different lighting conditions to simulate up to a 1000 sensor nodes. For each data trace, a new trace is generated by randomly adjusting both the light intensity by $\pm$30\% and shifting the sampling time by $\pm$3 hours. For each of the 5 light data traces, we build 200 new data traces.
The 5 light data traces that we collected are already covering a variety of indoor lighting conditions (i.e. Window, Door, Middle of Office, Conference Room and Stair Access), so the new nodes are placed on similar conditions but will be subject to a lower or higher light intensity based on the distance from the light source. As light-patterns are human activity dependent, shifting light availability by time captures variations in activities. With the 1000 simulated data traces, we use the first week of light data and calculate the mean of the light intensity for each node. We then group all the nodes in 5 different clusters based on mean light intensity and build a Q-Table (i.e. \textit{Cluster Q-Table}) for each of the 5 clusters. 

\begin{table}[ht]
\centering
  \caption{Total Rewards achieved for each lighting condition on a Q-Table generated for 1000 nodes and a Q-Table generated after clustering all the data-light traces}
  \vspace{-2mm}
  \label{tab:1000_res}
  \begin{tabular}{c c c c c c}
    \toprule
    Total & Window & Middle & Door & Confer & Stair\\
   Reward &  & Office &  & Room & Access\\
    \midrule
    1000 Nodes & \multirow{2}{0.6cm}{1290} & \multirow{2}{0.5cm}{420} & \multirow{2}{0.5cm}{369} & \multirow{2}{0.6cm}{1527} & \multirow{2}{0.5cm}{238} \\
    Q-Table &  & &  &  &  \\\hline
    
    Cluster & \multirow{2}{0.6cm}{1472} & \multirow{2}{0.5cm}{988} & \multirow{2}{0.5cm}{390} & \multirow{2}{0.6cm}{1580} &  \multirow{2}{0.5cm}{475}\\
    Q-Table&  &  &  &  &  \\
    
  \bottomrule
  \vspace{-1em}
\end{tabular}
\end{table}

Table \ref{tab:1000_res} reports the total reward obtained by running the original 5 data traces on \textit{(i)} a single general Q-Table built by using all the 1000 nodes light traces (i.e. \textit{1000 nodes Q-Table}), and \textit{(ii)} on the \textit{Cluster Q-Table}.
The Q-Tables generated after clustering the nodes (i.e. \textit{Cluster Q-Table}), are outperforming the Q-Table built by using the 1000 nodes. In the Middle of an Office case, the clustered Q-Table can achieve up to 235\% more reward compared to the 1000 nodes Q-Table.
Those results indicate that the \textit{Cluster Q-Table} is able to store individual characteristic of the 200 clustered nodes and to use those information to maximize the performance of the nodes. On the other hand, the \textit{1000 Nodes Q-Table} sacrifices maximum performance to allow the management of all the 1000 nodes. We leave automated clustering of nodes based on lighting characteristics to future work.


\subsection{Transfer Learning Results}
\label{transf_lern_result}

\begin{figure}[th]
	\centering
	\includegraphics[width=0.9\linewidth]{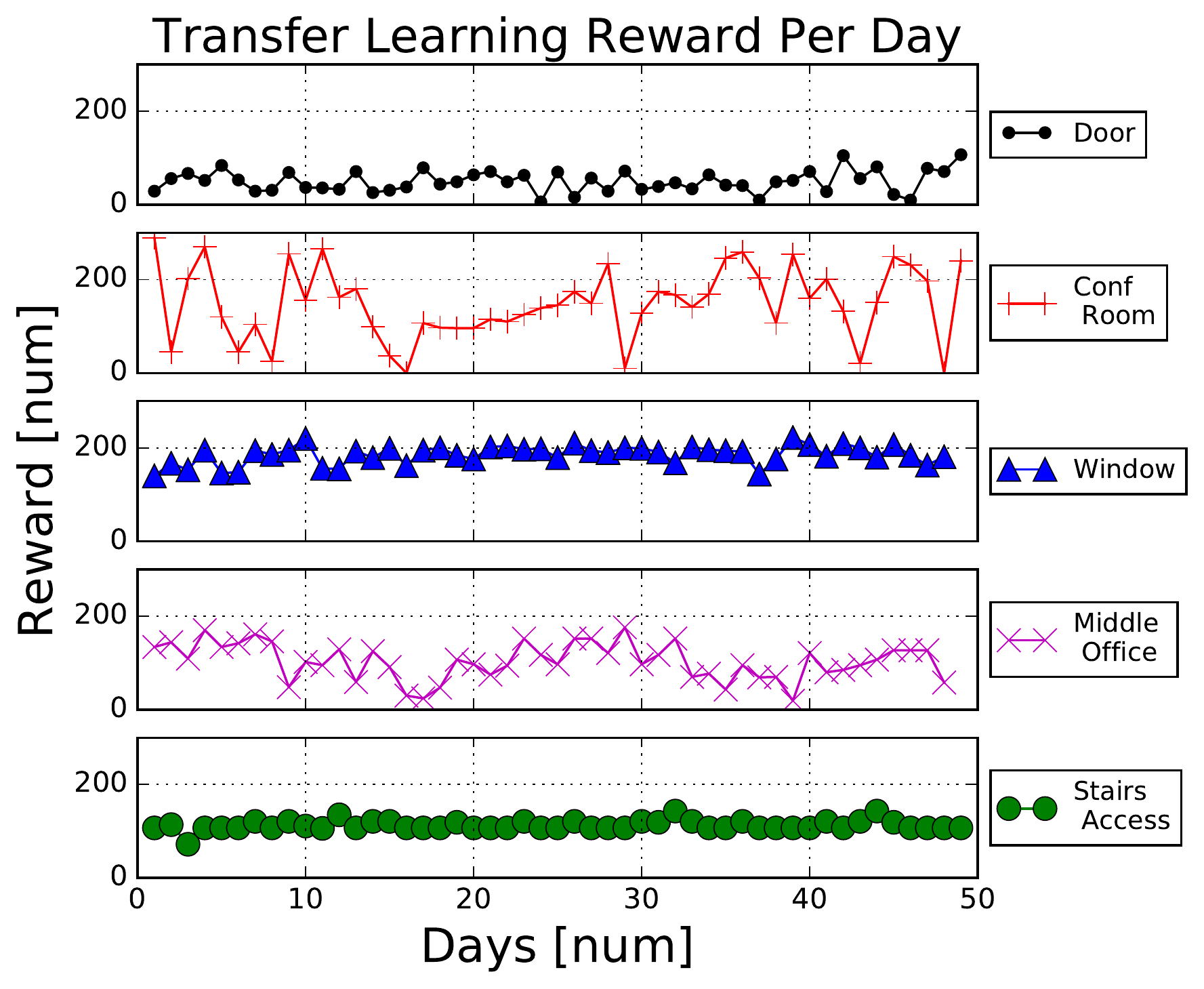}
     \vspace{-5mm}
	\caption{Rewards Obtained with Transfer Learning}
    \label{fig:transf_learn}
    \vspace{-1mm}
\end{figure}

We use the first week of light data for all the 5 different lighting conditions to train a general Q-Table. After convergence, we use this Q-Table to start the execution of the Day-by-Day Learning experiment instead of starting from an empty Q-Table. Results are reported in Figure \ref{fig:transf_learn}.

\vspace{-2mm}

All the 5 nodes are always collecting positive rewards even after the firsts days. Compared to the \textit{Day-by-Day Learning} experiment this is a great result, since almost all the nodes were achieving storage energy depletion as reported from Table \ref{tab:dynamic_res}.
This confirms that the information extracted from of a pre-calculated Q-Table built using general lightning trends, can be used to speed up the learning process of learning for different lighting conditions.

\section{Conclusion and Future Work}
In this work, we proposed and apply several solutions to scale the configuration of energy harvesting sensor with reinforcement learning. An adaptive on-policy RL solution that reduces the training phase after deployment has been tested. Results show that nodes can effectively adapt their sensing rate to different lighting conditions without depleting their stored energy and while reducing the number of on-policy training to up 81\% compared to a standard policy that runs on-line policy training every day. We also show that transfer RL can reduce the training phase, making the nodes operational within the first day. Finally, prior solutions consider one RL policy for each sensor node and affects the scalability. We show that the use of a single policy for sensors that share similar lighting conditions can still effectively configure the sensor nodes. 
We focus on simulation as a proof of concept in this work and will perform real experiments in future work.

\section*{Acknowledgments}

This work is supported by the National Science Foundation grant 
BD Spokes 1636879

\bibliographystyle{ACM-Reference-Format}
\bibliography{Biblio2}

\end{document}